\newcommand*{\COLT}{}

\newcommand*{\CAMREADY}{}

\ifdefined\NIPS
	\PassOptionsToPackage{numbers}{natbib}
	\documentclass{article}
	\ifdefined\CAMREADY	
		\usepackage[final]{nips_2016}
	\else
		\usepackage{nips_2016}
	\fi
	\usepackage[utf8]{inputenc} 
	\usepackage[T1]{fontenc}    
	\usepackage{hyperref}       	
	\usepackage{url}            	
	\usepackage{booktabs}       
	\usepackage{amsfonts}       
	\usepackage{nicefrac}       	
	\usepackage{microtype}      
\fi
\ifdefined\CVPR
	\documentclass[10pt,twocolumn,letterpaper]{article}
	\usepackage{cvpr}
	\usepackage{times}
	\usepackage{epsfig}
	\usepackage{graphicx}
	\usepackage{amsmath}
	\usepackage{amssymb}
	\usepackage[square,comma,numbers]{natbib}
\fi
\ifdefined\AISTATS
	\documentclass[twoside]{article}
	\ifdefined\CAMREADY
		\usepackage[accepted]{aistats2015}
	\else
		\usepackage{aistats2015}
	\fi
	\usepackage{url}
	\usepackage[round,authoryear]{natbib}
\fi
\ifdefined\ICML
	\documentclass{article}
	\usepackage{times}
	\usepackage{graphicx}
	\usepackage{subfigure}
	\usepackage{algorithm}
	\usepackage{algorithmic}
	\usepackage{hyperref}
	
	\ifdefined\CAMREADY
		\usepackage[accepted]{icml2016}
	\else
		\usepackage{icml2016}
	\fi
	\usepackage{natbib}
\fi
\ifdefined\ICLR
	\documentclass{article}
	\usepackage{iclr2017_conference,times}
	\usepackage{hyperref}
	\usepackage{url}

	\ifdefined\CAMREADY
		\iclrfinalcopy
	\fi
\fi
\ifdefined\COLT
	\ifdefined\CAMREADY
		\documentclass{colt2017}
	\else
		\documentclass[anon,12pt]{colt2017}
	\fi
	\usepackage{times}
\fi

\usepackage{color}
\usepackage{amsfonts}
\usepackage{algorithm}
\usepackage{algorithmic}
\usepackage{graphicx}
\usepackage{nicefrac}
\usepackage{bbm}
\usepackage{wrapfig}
\usepackage{tikz}
\usepackage[titletoc,title]{appendix}
\usepackage{stmaryrd}
\usepackage{enumerate}
\ifdefined\COLT
\else
	\usepackage{amsthm}
\fi

\ifdefined\COLT
	\newtheorem{claim}[theorem]{Claim}
	\newtheorem{fact}[theorem]{Fact}
	\newtheorem{question}{Question}
	
\else
	\newtheorem{definition}{Definition}
	
	\newtheorem{corollary}{Corollary}
	\newtheorem{theorem}{Theorem}

	\newtheorem{example}{Example}
	
	\newtheorem{claim}{Claim}
	
	\newtheorem{question}{Question}
\fi

\def\be{\begin{equation}}
\def\ee{\end{equation}}
\def\beas{\begin{eqnarray*}}
\def\eeas{\end{eqnarray*}}
\def\bea{\begin{eqnarray}}
\def\eea{\end{eqnarray}}

\newcommand{\aaa}{{\mathbf a}}

\newcommand{\A}{{\mathcal A}}
\newcommand{\B}{{\mathcal B}}
\newcommand{\CC}{{\mathcal C}}

\newcommand{\HH}{{\mathcal H}}

\newcommand{\M}{{\mathcal M}}
\newcommand{\NN}{{\mathcal N}}

\newcommand{\T}{{\mathcal T}}

\newcommand{\OO}{{\mathcal O}}

\newcommand{\R}{{\mathbb R}}
\newcommand{\N}{{\mathbb N}}

\DeclareMathOperator*{\mean}{mean}

\newcommand{\mat}[1]{\llbracket#1\rrbracket}

\newcommand{\otimesg}{\otimes_g}
\newcommand{\code}[1]{\texttt{#1}}

\newcommand{\deepsimnets}{cohen2016deep}
\newcommand{\cactd}{cohen2016expressive}
\newcommand{\crngtd}{cohen2016convolutional}
\newcommand{\inductive}{cohen2017inductive}
\newcommand{\tmm}{sharir2017tensorial}
\newcommand{\dcnmtd}{cohen2017boosting}
\newcommand{\overlaps}{sharir2017expressive}
\newcommand{\dlquantum}{levine2017deep}

\ifdefined\CVPR
	\usepackage[breaklinks=true,bookmarks=false]{hyperref}
	\ifdefined\CAMREADY
		\cvprfinalcopy 
	\fi
	
\fi

\ifdefined\ICML
	\newcommand*{\ABBR}{}
\fi
\ifdefined\NIPS
	\newcommand*{\ABBR}{}
\fi
\ifdefined\ICLR
	\newcommand*{\ABBR}{}
\fi
\ifdefined\COLT
	\newcommand*{\ABBR}{}
\fi
\ifdefined\ABBR
	\newcommand{\eg}{\emph{e.g.}}
	\newcommand{\ie}{\emph{i.e.}}
	\newcommand{\cf}{\emph{cf.}}
	\newcommand{\etc}{\emph{etc.}}
	
	\newcommand{\wrt}{with respect to}

\fi

\begin{document}

\ifdefined\NIPS
	\title{Analysis and Design of Convolutional Networks via Hierarchical Tensor Decompositions}
	\author{
	Nadav Cohen \\
	The Hebrew University of Jerusalem \\
	\texttt{cohennadav@cs.huji.ac.il} \\
	\And 
	Or Sharir \\
	The Hebrew University of Jerusalem \\
	\texttt{or.sharir@cs.huji.ac.il} \\	
	\And 
	Yoav Levine \\
	The Hebrew University of Jerusalem \\
	\texttt{yoavlevine@cs.huji.ac.il} \\	
	\And	
	Ronen Tamari \\
	The Hebrew University of Jerusalem \\
	\texttt{ronent@cs.huji.ac.il} \\
	\And
	David Yakira \\
	The Hebrew University of Jerusalem \\
	\texttt{davidyakira@cs.huji.ac.il} \\	
	\And	
	Amnon Shashua \\
	The Hebrew University of Jerusalem \\
	\texttt{shashua@cs.huji.ac.il} \\
	}
	\maketitle
\fi
\ifdefined\CVPR
	\title{Analysis and Design of Convolutional Networks via Hierarchical Tensor Decompositions}
	\author{
	Nadav Cohen \\
	The Hebrew University of Jerusalem \\	
	\texttt{cohennadav@cs.huji.ac.il} \\
	\and
	Or Sharir \\
	The Hebrew University of Jerusalem \\
	\texttt{or.sharir@cs.huji.ac.il} \\	
	\and
	Yoav Levine \\
	The Hebrew University of Jerusalem \\
	\texttt{yoavlevine@cs.huji.ac.il} \\	
	\and
	Ronen Tamari \\
	The Hebrew University of Jerusalem \\
	\texttt{ronent@cs.huji.ac.il} \\	
	\and
	David Yakira \\
	The Hebrew University of Jerusalem \\
	\texttt{davidyakira@cs.huji.ac.il} \\	
	\and
	Amnon Shashua \\
	The Hebrew University of Jerusalem \\
	\texttt{shashua@cs.huji.ac.il} \\
	}
	\maketitle
\fi
\ifdefined\AISTATS
	\twocolumn[
	\aistatstitle{Analysis and Design of Convolutional Networks via Hierarchical Tensor Decompositions}
	\ifdefined\CAMREADY
		\aistatsauthor{Nadav Cohen \And Or Sharir \And Yoav Levine \And Ronen Tamari \And David Yakira \And Amnon Shashua}
		\aistatsaddress{The Hebrew University of Jerusalem \And The Hebrew University of Jerusalem \And The Hebrew University of Jerusalem \And The Hebrew University of Jerusalem \And The Hebrew University of Jerusalem \And The Hebrew University of Jerusalem}
	\else
		\aistatsauthor{Anonymous Author 1 \And Anonymous Author 2 \And Anonymous Author 3}
		\aistatsaddress{Unknown Institution 1 \And Unknown Institution 2 \And Unknown Institution 3}
	\fi
	]	
\fi
\ifdefined\ICML
	\icmltitlerunning{Analysis and Design of Convolutional Networks via Hierarchical Tensor Decompositions}
	\twocolumn[
	\icmltitle{Analysis and Design of Convolutional Networks via Hierarchical Tensor Decompositions}
	\icmlauthor{Nadav Cohen}{cohennadav@cs.huji.ac.il}
	\icmladdress{The Hebrew University of Jerusalem}
	\icmlauthor{Or Sharir}{or.sharir@cs.huji.ac.il}
	\icmladdress{The Hebrew University of Jerusalem}
	\icmlauthor{Yoav Levine}{yoavlevine@cs.huji.ac.il}
	\icmladdress{The Hebrew University of Jerusalem}
	\icmlauthor{Ronen Tamari}{ronent@cs.huji.ac.il}
	\icmladdress{The Hebrew University of Jerusalem}
	\icmlauthor{David Yakira}{davidyakira@cs.huji.ac.il}
	\icmladdress{The Hebrew University of Jerusalem}	
	\icmlauthor{Amnon Shashua}{shashua@cs.huji.ac.il}
	\icmladdress{The Hebrew University of Jerusalem}
	\icmlkeywords{Convolutional Networks, Expressiveness, Hierarchical Tensor Decompositions}
	\vskip 0.3in
	]
\fi
\ifdefined\ICLR
	\title{Analysis and Design of Convolutional Networks via Hierarchical Tensor Decompositions}
	\author{Nadav Cohen \& Or Sharir \& Yoav Levine \& Ronen Tamari \& David Yakira \& Amnon Shashua\\ 
	\texttt{\{cohennadav,or.sharir,yoavlevine,ronent,davidyakira,shashua\}@cs.huji.ac.il}}
	\maketitle
\fi
\ifdefined\COLT
	\title[Analysis and Design of Convolutional Networks via Hierarchical Tensor Decompositions]{Analysis and Design of Convolutional Networks \\ via Hierarchical Tensor Decompositions}
	\coltauthor{
	\Name{Nadav Cohen} \Email{cohennadav@cs.huji.ac.il}\\
	\Name{Or Sharir} \Email{or.sharir@cs.huji.ac.il}\\
	\Name{Yoav Levine} \Email{yoavlevine@cs.huji.ac.il}\\
	\Name{Ronen Tamari} \Email{ronent@cs.huji.ac.il}\\
	\Name{David Yakira} \Email{davidyakira@cs.huji.ac.il}\\
	\Name{Amnon Shashua} \Email{shashua@cs.huji.ac.il}\\
	\addr The Hebrew University of Jerusalem}
	\maketitle
\fi

\begin{abstract}

The driving force behind convolutional networks~--~the most successful deep learning architecture to date, is their expressive power.
Despite its wide acceptance and vast empirical evidence, formal analyses supporting this belief are scarce. 
The primary notions for formally reasoning about expressiveness are efficiency and inductive bias. 
Expressive efficiency refers to the ability of a network architecture to realize functions that require an alternative architecture to be much larger. 
Inductive bias refers to the prioritization of some functions over others given prior knowledge regarding a task at hand. 
In this paper we overview a series of works written by the authors, that through an equivalence to hierarchical tensor decompositions, analyze the expressive efficiency and inductive bias of various convolutional network architectural features (depth, width, strides and more). 
The results presented shed light on the demonstrated effectiveness of convolutional networks, and in addition, provide new tools for network design.
\footnote{This work was supported by the Intel Collaborative Research Institute for Computational Intelligence (ICRI-CI), and is part of the ``Why \& When Deep Learning works~--~looking inside Deep Learning'' ICRI-CI paper bundle.}
\end{abstract}

\ifdefined\COLT
	\medskip
\fi

\section{Introduction} \label{sec:intro}

\emph{Convolutional networks}~(\cite{lecun1995convolutional}) are the cornerstone of modern deep learning.
Since the work of~\cite{Krizhevsky:2012wl}, they prevail in the domain of visual recognition, and recently, they have also been delivering state of the art results in speech and text processing tasks (see for example~\cite{van2016wavenet,kalchbrenner2016neural}).
As opposed to classic deep network architectures, such as the multilayer perceptron (feed-forward fully-connected neural network~--~\cite{rosenblatt1961principles}), employing a modern convolutional network involves setting dozens or even hundreds of architectural parameters.
Namely, besides the basic choices of network depth, width of each layer, and type of non-linear activations (\eg~sigmoid or~ReLU~--~\cite{nair2010rectified}), one must decide on the type of pooling operator in each layer (\eg~max or average), the kernel sizes and strides in every convolution/pooling, the connectivity scheme to employ (\eg~skip connections~--~\cite{he2015deep}), and much more.
To date, these architectural choices are typically made heuristically, based on past experience, conventional wisdom, and trial-and-error.
This often leads to lengthy and inefficient development cycles, ultimately concluding in suboptimal results.
More principled design practices, which could be made available by a more formal understanding of modern convolutional network architectures, are thus of great interest.

It is widely accepted that the driving force behind convolutional networks, and deep networks in general, is their \emph{expressiveness}, \ie~their ability to compactly represent rich and effective classes of functions.
The primary notions for formally reasoning about expressiveness are \emph{efficiency} and \emph{inductive bias}.
Efficiency refers to a situation where one network must grow unfeasibly large in order to realize (or approximate) functions of another.
Inductive bias refers to real-world tasks requiring specific types of functions (\eg~translation invariant), not just arbitrary ones.
Our interest lies on the expressiveness of convolutional networks, or more specifically, on the expressive efficiency and inductive bias brought forth by their various architectural features.
We review in this paper a series of works written by the authors (\cite{\cactd,\deepsimnets,\tmm,\crngtd,\inductive,\overlaps,\dcnmtd,\dlquantum}), which address these topics through the mathematical notion of \emph{hierarchical tensor decompositions} (see~\cite{Hackbusch-book} for a comprehensive introduction).
Our presentation here is soft and oftentimes informal, with the objective of creating a manuscript accessible to a wide range of audience.
For an exact and formal presentation, we refer the reader to the papers we review.

\section{Expressive Efficiency and Inductive Bias} \label{sec:eff_ind}

\begin{figure*}
\includegraphics[width=\textwidth]{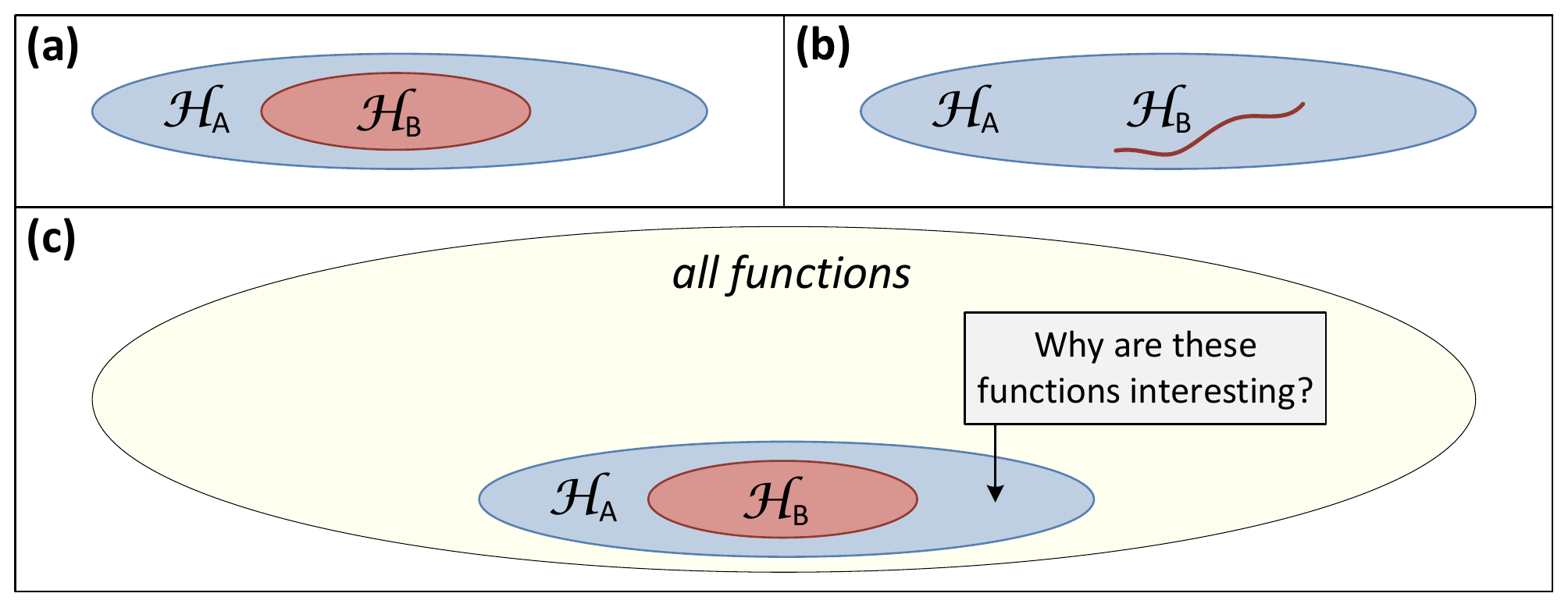}
\vspace{-7mm}
\caption{
Illustrations of expressive efficiency and inductive bias.
$\HH_A$~and~$\HH_B$ are the sets of functions practically realizable by network architectures~$A$ and~$B$ (respectively).
\textit{\textbf{(a)}}~Expressive efficiency~--~$\HH_A$ is a strict superset of~$\HH_B$.
\textit{\textbf{(b)}}~Complete expressive efficiency~--~$\HH_B$ has zero ``volume'' inside~$\HH_A$.
\textit{\textbf{(c)}}~Inductive bias~--~efficiency of~$A$ over~$B$ does not imply that~$\HH_A$ contains meaningful functions; this can only be explained through the requirements of real-world problems.
}
\label{fig:eff_ind}
\vspace{-5mm}
\end{figure*}

As stated in the introduction, \emph{expressive efficiency} refers to a situation where one network must grow unfeasibly large in order to realize (or approximate) functions of another.
More explicitly, consider network architectures~$A$ and~$B$, with size parameters~$r_A$ and~$r_B$ (respectively).
For example, $A$~could be an instance of AlexNet~(\cite{Krizhevsky:2012wl}) with each layer (convolutional and fully-connected) comprising~$r_A$ channels, whereas~$B$ could be an instance of ResNet~(\cite{he2015deep}) with~$r_B$ channels in each of its layers.
Denote by~$\HH_A$ (respectively~$\HH_B$) the set of functions that may be realized by~$A$ (respectively~$B$) with size parameter~$r_A$ (respectively~$r_B$) that is small enough for practical implementation.
We say that architecture~$A$ is efficient \wrt~architecture~$B$ if~$\HH_A$ is a strict superset of~$\HH_B$ (see illustration in fig.~\ref{fig:eff_ind}(a)), meaning that~$A$ can realize with practical size anything that~$B$ can, whereas the converse does not hold~--~there exist functions compactly realizable by~$A$ that cannot be practically replicated by~$B$.

Given that~$A$ is efficient \wrt~$B$, a natural question arises:
How many of the functions realizable by~$A$ reflect its efficiency over~$B$?
Is it just one function that~$A$ can realize compactly and~$B$ cannot, or are there many?
This question amounts to reasoning about the ``volume'' of~$\HH_B$ inside~$\HH_A$~--~if the volume is small, a significant portion of the functions compactly realizable by~$A$ lay outside the reach of~$B$, whereas on the other hand, a large volume implies that~$B$ comes close to~$A$ in terms of the functions it supports.
The strongest form of efficiency, referred to as \emph{complete}, takes place when the volume of~$\HH_B$ in~$\HH_A$ is essentially zero (see illustration in fig.~\ref{fig:eff_ind}(b)).
In this case, almost all functions realizable by~$A$ cannot be replicated by~$B$ unless that is unfeasibly large.

For concreteness, we provide below definitions of expressive efficiency and completeness that are slightly more formal than the above:
\vspace{-2mm}
\begin{definition} \label{def:eff}
Let~$A$ and~$B$ be network architectures with size parameters~$r_A$ and~$r_B$ (respectively).
We say that~$A$ is \emph{expressively efficient} \wrt~$B$ if the following two conditions hold:
\vspace{-2mm}
\begin{enumerate}[(i)]
\item Any function realized by~$B$ with size~$r_B$ can be replicated by~$A$ with size~$r_A$ that is no more than linear in~$r_B$ (\ie~$r_A\in\OO(r_B)$).
\vspace{-2mm}
\item There exist functions realized by~$A$ with size~$r_A$ that cannot be replicated by~$B$ unless its size~$r_B$ is super-linear in~$r_A$ (\ie~$r_B\in\Omega(f(r_A))$ for a super-linear function~$f(\cdot)$)
\end{enumerate}
\vspace{-2mm}
The expressive efficiency of~$A$ over~$B$ is \emph{complete} if randomizing the weights of~$A$ by any continuous distribution leads, with probability~$1$, to a function satisfying condition~(ii).
\end{definition}

\medskip

Expressive efficiency alone does not convey the entire story behind the effectiveness of functions realized by deep networks.
Under any network architecture, the set of functions realizable with practical size is merely a small fraction of all possible functions.
Accordingly, even if architecture~$A$ is expressively efficient \wrt~architecture~$B$, we have no information indicating that functions compactly realizable by~$A$ would be effective in practice.
In other words, even if~$\HH_A$ is a strict superset of~$\HH_B$, it is still a mere corner in the space of all functions, and a-priori, may not include any meaningful function (see illustration in fig.~\ref{fig:eff_ind}(c)).
To understand why for certain architectures, \eg~convolutional networks, $\HH_A$ is so effective in practice, one must consider the \emph{inductive bias}, \ie~the actual needs of real-world problems.
Functions required for successful execution of tasks such as image classification or speech-to-text annotation, are not arbitrary~--~there are certain task-dependent properties, for example smoothness or translation invariance, that must be met.
Accordingly, a given network architecture need not realize all functions, only those possessing certain properties.
Empirical evidence suggests that by properly designing a convolutional network, functions fulfilling requirements of various tasks become available.
Formal understanding of this phenomenon is lacking.

\subsection{Questions on the Expressive Efficiency and Inductive Bias of Convolutional Networks} \label{sec:eff_ind:qs}

Our interest lies on the expressive efficiency and inductive bias brought forth by the various architectural features of modern convolutional networks.
Below are the specific questions we address.

\begin{question}[Efficiency of Depth~--~addressed in sec.~\ref{sec:eff_depth}]
\label{question:eff_depth}
\normalfont
Perhaps the most prominent empirical finding of deep learning, which in some sense identifies the field, is that deep networks, when operated appropriately, greatly outperform shallow ones (see~\cite{LeCun:2015dt} for a survey of such results).
The conventional argument for explaining this phenomenon is that depth brings forth a representational power that is otherwise unattainable.
Formally, it amounts to saying that deep networks are expressively efficient \wrt~shallow ones.
This proposition, which traces back to classical questions from the world of circuit complexity, has recently been proven for various network architectures (see for example~\cite{bengio2011shallow,pascanu2013number,montufar2014number,telgarsky2015representation,eldan2015power,poggio2015theory,mhaskar2016learning}).
For convolutional networks however, the proposition has not been proven, and the question of whether or not depth brings forth expressive efficiency remains open.
Moreover, even if one makes the reasonable assumption by which convolutional networks, similarly to other architectures, admit depth efficiency, it still is unclear how frequent the latter is, and in particular, whether or not it is complete.
Completeness of depth efficiency has never been established, for any network architecture of a practical nature.
\end{question}

\begin{question}[Inductive Bias of Convolution/Pooling Geometry~--~addressed in sec.~\ref{sec:ind_geo}]
\label{question:ind_geo}
\normalfont
A key ingredient of convolutional networks is the locality of their convolution and pooling (decimation) operations.
Traditionally, convolution and pooling windows are chosen to be contiguous blocks (squares in~2D networks, intervals in~1D), reflecting an intuitive assumption by which such geometries are suitable for data of a continuous nature (\eg~images or audio).
Recently however, several works have demonstrated that different geometries, such as non-contiguous windows with internal dilations (\cf~\cite{yu2015multi,van2016wavenet,kalchbrenner2016neural}), or even windows with dynamically learned shapes (\cf~\cite{filter2017li}), can lead to improved performance.
We would like to understand the relations between a network's convolution/pooling geometry, the set of functions it can model, and the suitability of this set for different tasks.
That is to say, we would like to understand the inductive bias governing geometries of convolution and pooling windows in convolutional networks.
Formal results on this line are not only of theoretical interest~--~they may potentially provide practical guidelines for tailoring a network's convolution/pooling geometry in accordance with a task at hand.
\end{question}

\begin{question}[Efficiency of Overlapping Operations~--~addressed in sec.~\ref{sec:eff_overlap}]
\label{question:eff_overlap}
\normalfont
Modern convolutional networks, \eg~VGG~(\cite{simonyan2014very}) or GoogLeNet~(\cite{Szegedy:2014tb}), include a mix of convolution and pooling operations, some of which are overlapping (stride smaller than window size), while others are not (stride and window size equal).
Empirical evidence suggests that non-overlapping operations are beneficial, but nonetheless, must be accompanied by overlapping operations in order to produce competitive performance.
We would like to understand whether this need for overlaps can be attributed to expressiveness, or more specifically, whether convolutional networks with overlapping operations are expressively efficient \wrt~ones without.
\end{question}

\begin{question}[Inductive Bias of Layer Widths~--~addressed in sec.~\ref{sec:ind_width}]
\label{question:ind_width}
\normalfont
A fundamental architectural choice to be made when designing a convolutional network is the width of (number of channels in) each layer.
At present, there are no firm principles for making this decision~--~in the majority of cases layer widths are either set uniformly across a network, or such that deeper layers are wider, so as to avoid ``representational bottlenecks''~(\cite{szegedy2016rethinking}).
Given a fixed amount of computational resources, it is unclear what would be an effective distribution of layer widths across a network, and how this depends on the particular task at hand.
From a representational perspective, this boils down to reasoning about the inductive bias of layer widths, \ie,~about the implication of widening one layer versus another in terms of the functions a network can realize.
A formal treatment of this question could pave the way to more principled convolutional network designs, in which layer widths are tailored to the nature of a given task.
\end{question}

\begin{question}[Efficiency of Connectivity~--~addressed in sec.~\ref{sec:eff_conn}]
\label{question:eff_conn}
\normalfont
The classic convolutional network architecture, oftentimes referred to as LeNet (see~\cite{lecun1995convolutional}), consists of layers concatenated one after the other in a feed-forward (chain) scheme.
Until recently, networks adhering to this architectural paradigm provided state of the art visual recognition performance.
In~2014, with the rise of GoogLeNet~(\cite{Szegedy:2014tb}), a new type of convolutional networks has emerged.
These networks no longer follow the simple feed-forward approach, but rather run layers in parallel, employing various connectivity (split/merge) schemes.
In~2015, connectivity schemes in convolutional networks took one step further, with the introduction of ResNet~(\cite{he2015deep}), whose layers are linked through ``skip connections''.
Nowadays, nearly all state of the art convolutional networks (\eg~\cite{huang2016densely,van2016wavenet,kalchbrenner2016neural}), for visual recognition as well as audio and text processing tasks, employ elaborate connectivity schemes.
The question we ask is whether this can be understood in terms of expressive efficiency, \ie,~whether connectivity schemes bear the potential to create networks that are expressively efficient \wrt~the standard feed-forward architecture.
\end{question}

\section{Convolutional Arithmetic Circuits and Hierarchical Tensor Decompositions} \label{sec:cac_htd}

To analyze the expressive efficiency and inductive bias (see sec.~\ref{sec:eff_ind}) of convolutional networks, and in particular, to address the questions laid out in sec.~\ref{sec:eff_ind:qs}, we focus on a family of models named \emph{convolutional arithmetic circuits}.
Convolutional arithmetic circuits are convolutional networks with a particular choice of non-linearities.
Namely, they arise by setting point-wise activations to be linear (as opposed to sigmoid or ReLU), and pooling operators to be based on product (as opposed to max or average).\footnote{
As an alternative viewpoint, convolutional arithmetic circuits can be seen as sum-product networks~(\cite{Poon-Domingos2011}) whose structure is convolutional.
}
The reason we focus on convolutional arithmetic circuits is their intimate relation to various mathematical fields (tensor analysis, measure theory, functional analysis, theoretical physics, graph theory and more), rendering them especially amendable to theoretical analyses.
We will see in sec.~\ref{sec:eff_depth:crn} how mathematical machinery developed for the analysis of convolutional arithmetic circuits can be adapted to account for other types of convolutional networks as well, for example ones with ReLU activation and max or average pooling.
We note that besides their theoretical merits, convolutional arithmetic circuits also deliver promising results in practice.
Specifically, they excel in computationally constrained settings~(\cite{\deepsimnets}), and give state of the art results in classification under missing data~(\cite{\tmm}).\footnote{
An implementation of convolutional arithmetic circuits (also known as \emph{SimNets}) for Caffe toolbox~(\cite{jia2014caffe}) can be found on-line at \url{https://github.com/HUJI-Deep/caffe-simnets}.
}

\medskip

\begin{figure*}
\includegraphics[width=\textwidth]{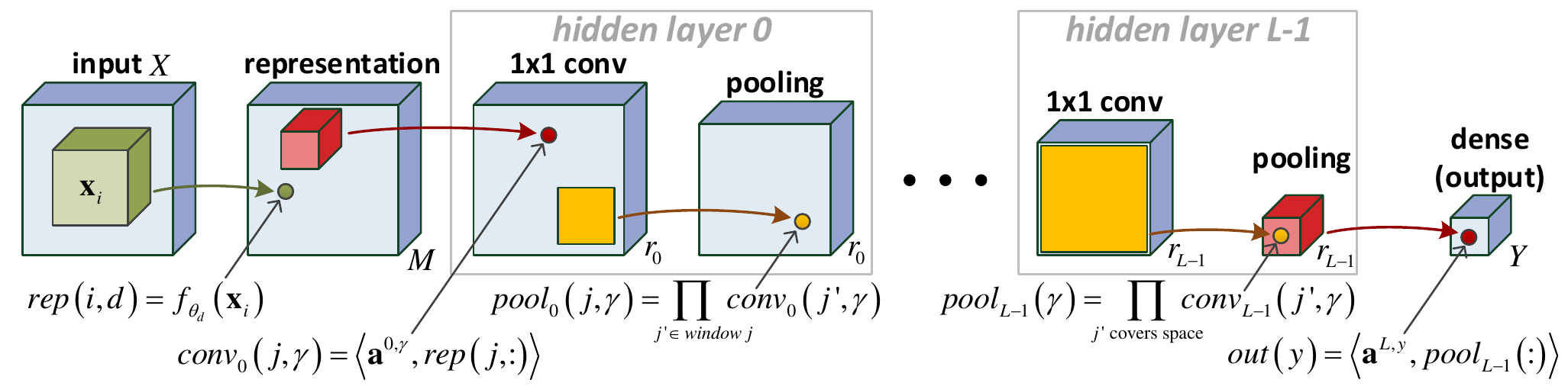}
\vspace{-7mm}
\caption{
Baseline convolutional arithmetic circuit architecture~--~the starting point of our analyses.
}
\label{fig:base_cac}
\vspace{-5mm}
\end{figure*}

The convolutional arithmetic circuit architecture we consider as baseline is the one depicted in fig.~\ref{fig:base_cac}.
It is a 2D convolutional network comprising an initial convolutional layer (referred to as \emph{representation}) followed by~$L$ hidden layers, which in turn are followed by a dense (linear) output layer.
Each hidden layer consists of a convolution and spatial pooling.
Besides the fact that the convolution's activations are linear ($\sigma(z){=}z$), and that the pooling operator is based on products ($P\{c_j\}{=}\prod_{j}c_j$), further restrictions of the architecture are that receptive fields of the convolution are~$1{\times}1$, and pooling windows do not overlap.
All of these limitations will be relieved as we move forward (sec.~\ref{sec:eff_depth:crn} and~\ref{sec:eff_overlap}).
At the starting point however, we remain with the baseline architecture (fig.~\ref{fig:base_cac}), as it holds strong connections to well-established mathematical constructions, in particular ones from the field of tensor analysis.

There are different ways to formulate a connection between convolutional networks and tensors (multi-dimensional arrays), the simplest being through the notion of grid tensors.
Let~$f(\cdot)$ be a real-valued function defined over the unit square in the plane, \ie~$f:[0,1]^2\to\R$.
For a positive integer~$M\in\N$, we may discretize the interval~$[0,1]$ into the~$M$ points $\{\nicefrac{1}{M},\nicefrac{2}{M},\ldots,\nicefrac{M}{M}\}$, and define an~$M{\times}M$ matrix holding function values over discretized inputs:
$$
\A:=
\begin{bmatrix}
f(\nicefrac{1}{M},\nicefrac{1}{M}) & f(\nicefrac{1}{M},\nicefrac{2}{M}) & \cdots & f(\nicefrac{1}{M},\nicefrac{M}{M}) \\
f(\nicefrac{2}{M},\nicefrac{1}{M}) & f(\nicefrac{2}{M},\nicefrac{2}{M}) & \cdots & f(\nicefrac{2}{M},\nicefrac{M}{M}) \\
\vdots & \vdots & \ddots & \vdots \\
f(\nicefrac{M}{M},\nicefrac{1}{M}) & f(\nicefrac{M}{M},\nicefrac{2}{M}) & \cdots & f(\nicefrac{M}{M},\nicefrac{M}{M})
\end{bmatrix}
$$
This matrix is in fact a lookup table representing~$f(\cdot)$, that becomes larger and more fine-grained as~$M$ grows.
Suppose now that~$f(\cdot)$ was defined over the $N$-dimensional unit hypercube, \ie~$f:[0,1]^N\to\R$.
In this case the lookup table~$\A$ would transform from a matrix into a tensor (multi-dimensional array), having $N$~modes (axes) of length~$M$ each.
We refer to~$\A$ as the \emph{grid tensor} of the function~$f(\cdot)$.
A convolutional network with fixed weights can be viewed (without loss of generality) as a function over the unit hypercube, where the dimension~$N$ is equal to the number of input elements (\eg~pixel values or audio samples).
We will study such functions through their corresponding grid tensors.

Grid tensors of convolutional networks are typically of very high order, \ie~have many modes.
For example, if a network processes gray-scale images of size~$100{\times}100$, $N$~--~the order of (number of modes in) its grid tensor, will be~$10^4$, meaning there are~$M^{10^4}$ entries in the tensor.
Such exponentially large tensors are obviously impractical to manipulate or store directly.
They may however be represented efficiently through algebraic constructions named \emph{tensor decompositions}.
Tensor decompositions are essentially parameterizations that allow representation of large tensors with a relatively small number of parameters.
In the special case of order-$2$ tensors, \ie~matrices, most types of tensor decompositions boil down to simply a low-rank matrix decomposition.
As an example of the latter, consider the space~$\R^{10^6{\times}10^6}$, \ie~the space of matrices with size~$10^6{\times}10^6$.
Elements of this space are too large to be stored directly in a typical personal computer.
However, if we are willing to limit ourselves to a subset of this space comprising matrices of low rank, a compact parameterization immediately emerges.
For instance, if matrices of rank~$5$ or less are sufficient, any element in our subset can be represented as a product of two matrices~--~one of size~$10^6{\times}5$, and the other of size~$5{\times}10^6$.
We thus obtain a representation of tensors (matrices) with $10^{12}$~entries using only $10^7$~parameters~--~a manageable number even for a low-end handheld device.

\begin{figure*}
\includegraphics[width=\textwidth]{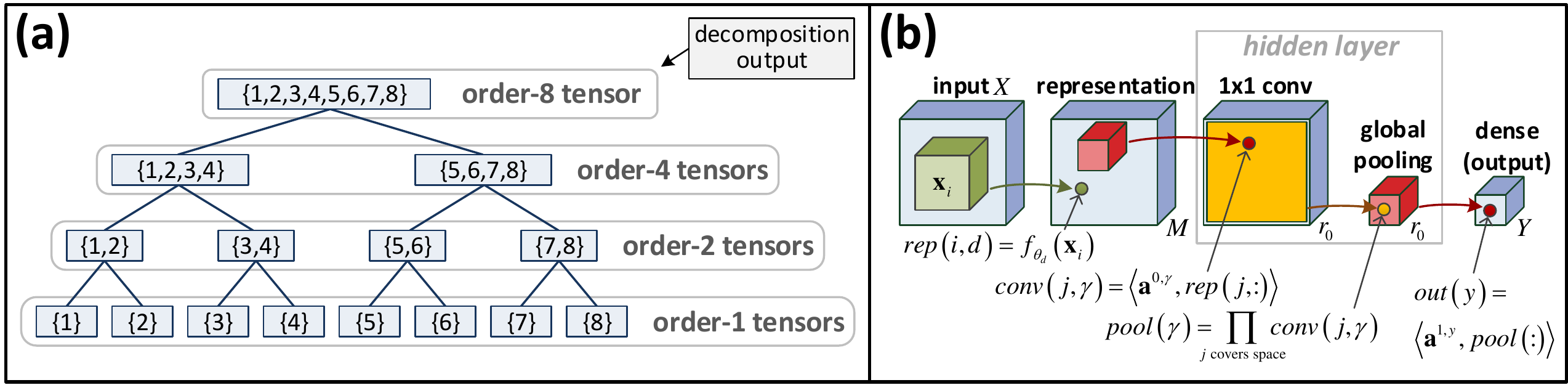}
\vspace{-7mm}
\caption{
\textit{\textbf{(a)}}~Illustration of a tree over tensor modes (axes).
Different mode trees give rise to different hierarchical tensor decompositions.
The decomposition corresponding to this tree operates in three stages: the first assembles vectors (order-$1$ tensors) into matrices (order-$2$ tensors), the second uses these matrices to create order-$4$ tensors, and the third (final stage) combines the latter into an order-$8$ tensor~--~decomposition output.
\textit{\textbf{(b)}}~Shallow (single hidden layer) convolutional arithmetic circuit.
}
\label{fig:mode_tree_shallow_cac}
\vspace{-5mm}
\end{figure*}

As opposed to the special case of matrices (order-$2$ tensors), decomposing tensors of a general order can be done in numerous ways.
A rich family of decompositions, which allows representing tensors of extremely high order, is the so-called hierarchical format, also known as \emph{hierarchical tensor decompositions}.
Introduced in~\cite{Hackbusch:2009jj} (and later generalized in~\cite{Hackbusch-book}), these decompositions represent tensors by incrementally constructing intermediate tensors of increasing order.
For example, suppose we are to decompose an order-$8$ tensor.
A hierarchical decomposition could operate in three stages: the first assembles vectors (order-$1$ tensors) into matrices (order-$2$ tensors), the second uses these matrices to construct order-$4$ tensors, and the third (final stage) combines the latter tensors into the final order-$8$ output.
This process can be described by a full binary tree\footnote{
A full binary tree is a tree in which all nodes but the leaves have exactly two children.
} over tensor modes, as illustrated in fig.~\ref{fig:mode_tree_shallow_cac}(a).
In general, when the order of a tensor is high, there are many possible trees over its modes, and each tree corresponds to a different hierarchical decomposition.

A key observation we make is that \emph{\textbf{convolutional arithmetic circuits are equivalent to hierarchical tensor decompositions}}.
More precisely, grid tensors of functions realized by the baseline convolutional arithmetic circuit architecture (fig.~\ref{fig:base_cac}) can be represented via hierarchical tensor decompositions.
The correspondence between networks and decompositions is bijective (one-to-one)~--~for every network structure (depth, width of each layer, geometry of pooling windows~\etc) there exists a unique decomposition that represents its grid tensors, and vice versa.
Under this correspondence, network weights (in convolution and output layers) are directly mapped to the parameters of the respective decomposition.
We present below two canonical examples of networks and their corresponding decompositions.
These examples will accompany us throughout the paper.

\begin{example}[Shallow Network $\longleftrightarrow$ CP Decomposition]
\label{example:shallow_cp}
\normalfont
Consider a shallow convolutional arithmetic circuit with a single hidden layer of width~$r_0$, as illustrated in fig.~\ref{fig:mode_tree_shallow_cac}(b).
The convolutional filters of this network are the vectors~$\{\aaa^{0,\gamma}\}_{\gamma=1}^{r_0}$, and the linear weights of output~$y$ are held in vector~$\aaa^{1,y}$.
Denote by~$\A^y$ the grid tensor of the function realized by output~$y$.
This tensor is given by the following formula:
\vspace{-1mm}
\be
\A^y=\sum\nolimits_{\gamma=1}^{r_0}a^{1,y}_{\gamma}\cdot\underbrace{\aaa^{0,\gamma}\otimes\aaa^{0,\gamma}\otimes\cdots\otimes\aaa^{0,\gamma}}_{\text{$N$ times}}
\label{eq:cp_decomp}
\vspace{-2mm}
\ee
where~$a^{1,y}_{\gamma}$ stands for coordinate~$\gamma$ of the vector~$\aaa^{1,y}$, and~$\otimes$ is simply the outer product operator.\footnote{
For example, if~$\B$ and~$\CC$ are order-$3$ and order-$4$ tensors ($3$-~and $4$-dimensional arrays) respectively, their outer product~$\B\otimes\CC$ is the order-$7$ tensor defined by: $(\B\otimes\CC)_{d_1{\ldots}d_7}=\B_{d_1{\ldots}d_3}\cdot\CC_{d_4{\ldots}d_7}$.
}
The formula in eq.~\ref{eq:cp_decomp} is an instance of the CANDECOMP/PARAFAC tensor decomposition, or \emph{CP decomposition} for short.
CP~decomposition can be viewed as a special case of a hierarchical decomposition, and is perhaps the most classic tensor decomposition recorded in the literature, dating back to the early 20th century (see~\cite{Kolda-Bader2009} for a historic survey).
\end{example}

\begin{example}[Deep Network $\longleftrightarrow$ HT Decomposition]
\label{example:deep_ht}
\normalfont
Consider now the deep convolutional arithmetic circuit obtained by setting each pooling window in the baseline architecture (fig.~\ref{fig:base_cac}) to have size~$2$.
Grid tensors of functions realized by this network are given by the Hierarchical Tucker decomposition, whose name we abbreviate as \emph{HT decomposition}.
HT decomposition was the first tensor decomposition to explicitly incorporate a hierarchical structure, based on a perfect binary mode tree\footnote{
A perfect binary tree is a tree in which all interior (non-leaf) nodes have exactly two children and all leaves have exactly the same depth.
}
as illustrated in fig.~\ref{fig:mode_tree_shallow_cac}(a).
Its introduction in~\cite{Hackbusch:2009jj} marks the dawn of hierarchical tensor decompositions as they are known today.
\end{example}

\medskip

To summarize this section, we presented an algebraic (sum-product) variant of convolutional networks named convolutional arithmetic circuits, and discussed its equivalence to hierarchical tensor decompositions.
There is a one-to-one correspondence between the structure of a network (depth, width of each layer~\etc) and the type of its respective decomposition, with network weights mapped to decomposition parameters.
This allows analyzing networks through their corresponding tensor decompositions, opening the door to a plurality of mathematical tools.
Hereafter, we make use of these tools to analyze the expressive efficiency and inductive bias (see sec.~\ref{sec:eff_ind}) of convolutional arithmetic circuits, as well as other types of convolutional networks (ones with ReLU activation and max or average pooling).

\section[Efficiency of Depth]{Efficiency of Depth \textit{(\cite{\cactd,\crngtd})}} \label{sec:eff_depth}

In this section we address question~\ref{question:eff_depth} in sec.~\ref{sec:eff_ind:qs}, dealing with the expressive efficiency brought forth by deepening convolutional networks.
As a first step in this direction, we compare the shallow and deep convolutional arithmetic circuits presented in sec.~\ref{sec:cac_htd} (examples~\ref{example:shallow_cp} and~\ref{example:deep_ht}), which correspond to CP and HT decompositions respectively.
Recall from def.~\ref{def:eff} that in order to establish expressive efficiency of the deep network \wrt~the shallow one, two propositions are to be proven:
\vspace{-2mm}
\begin{enumerate}[(i)]
\item Any function realized by the shallow network can be replicated by the deep network with no more than linear growth in size
\vspace{-2mm}
\item There exist functions realized by the deep network that cannot be replicated by the shallow network unless that is allowed to grow super-linearly
\end{enumerate}
\vspace{-2mm}
Proposition~(i) is trivial~--~it follows from the fact that the deep network reduces to the shallow one if we set all of its hidden convolutions but the first to be identity mappings.
Proposition~(ii) is much less obvious~--~we prove it by showing that under a matrix arrangement, ranks of grid tensors realized by the deep network are far greater than those brought forth by the shallow one.

\medskip

The process of arranging a tensor as a matrix is called matricization.
Let~$\A$ be a tensor with~$N$ modes (order-$N$), each of length~$M$.
Let~$(I,J)$ be a partition of these modes, \ie~$I$ and~$J$ are disjoint subsets of~$\{1,2,\ldots,N\}$ whose union covers the entire set.
The \emph{matricization of~$\A$ \wrt~$(I,J)$}, denoted~$\mat{\A}_{I,J}$, is an arrangement of~$\A$ as a matrix, with rows corresponding to modes indexed by~$I$, and columns corresponding to modes indexed by~$J$.
For example, suppose that~$N=5$, $I=\{2,3,5\}$ and~$J=\{1,4\}$.
In this case~$\mat{\A}_{I,J}$ is obtained by reordering the modes of~$\A$ via~$(2,3,5,1,4)$ (\eg~using NumPy's \code{transpose()} function or MATLAB's \code{permute()}), and then reshaping the resulting array to a~$M^3{\times}M^2$ matrix (\eg~with NumPy or MATLAB's \code{reshape()} functions).

The following claim, proven in~\cite{\inductive}, characterizes tensors generated by CP decomposition in terms of their ranks when subject to matricization:
\begin{claim} \label{claim:cp_ranks}
Tensors generated by CP decomposition, when matricized \wrt~any partition~$(I,J)$, have rank that does not exceed the number of terms (summands) in the decomposition.
\end{claim} 
Recall from example~\ref{example:shallow_cp} that in the CP decomposition corresponding to a shallow convolutional arithmetic circuit (eq.~\ref{eq:cp_decomp}), $r_0$~--~the number of terms, is precisely equal to the number of hidden channels in the network (see fig.~\ref{fig:mode_tree_shallow_cac}(b)).
Claim~\ref{claim:cp_ranks} thus implies that grid tensors of functions realized by the shallow convolutional arithmetic circuit have matricization ranks that do not exceed the number of hidden channels.

In stark contrast to CP decomposition, HT decomposition generates tensors with exponentially high matricization ranks.
This is formulated in the theorem below, proven in~\cite{\cactd}:
\begin{theorem} \label{theorem:ht_ranks_even_odd}
Almost every tensor generated by HT decomposition, when matricized \wrt~an even-odd partition ($I=\{1,3,\ldots\},~J=\{2,4,\ldots\}$), has an exponentially high rank.
\end{theorem}
From a network perspective, theorem~\ref{theorem:ht_ranks_even_odd} implies that grid tensors of functions realized by the deep convolutional arithmetic circuit (example~\ref{example:deep_ht}), when matricized \wrt~a particular (even-odd) partition, have ranks that are exponentially high.
Moreover, this holds for almost every grid tensor, meaning that if we randomize the weights of the deep network by some continuous distribution, with probability~$1$, we obtain a function whose grid tensor has an exponential matricization rank.

Taken together, claim~\ref{claim:cp_ranks} and theorem~\ref{theorem:ht_ranks_even_odd} lead to the following corollary:
\begin{corollary} \label{corollary:cac_eff_depth_complete}
Suppose we randomize the weights of a deep convolutional arithmetic circuit by some continuous distribution.
Then, with probability~$1$, we obtain functions that may only be replicated by a shallow convolutional arithmetic circuit if that has an exponential number of hidden channels.
\end{corollary}
Corollary~\ref{corollary:cac_eff_depth_complete} can be phrased succinctly by saying that \emph{\textbf{with convolutional arithmetic circuits, the expressive efficiency of depth is exponential and complete}}.
We have treated here the specific shallow and deep networks presented in sec.~\ref{sec:cac_htd} (examples~\ref{example:shallow_cp} and~\ref{example:deep_ht} respectively), but the methodology employed is readily applicable to arbitrary structures (instances of the baseline convolutional arithmetic circuit architecture~--~fig.~\ref{fig:base_cac}).
A less immediate step is the adaptation of our analysis to convolutional networks that are not arithmetic circuits, for example ones with ReLU activation and max or average pooling.
This is the topic of the subsection that follows.

\subsection{Convolutional Rectifier Networks} \label{sec:eff_depth:crn}

\emph{Convolutional rectifier networks} are convolutional networks with ReLU (Rectified Linear Unit~--~\cite{nair2010rectified}) activation and max or average pooling.
They are the most commonly used type of convolutional networks these days, and thus are of particular interest.
We demonstrate below how mathematical machinery developed for the analysis of convolutional arithmetic circuits can be adapted to account for convolutional rectifier networks as well.
Our use case will be the study of expressive efficiency brought forth by depth.

\medskip

Our analysis of convolutional arithmetic circuits is facilitated by their equivalence to hierarchical tensor decompositions (sec.~\ref{sec:cac_htd}).
The central operator in hierarchical tensor decompositions is the outer product~$\otimes$, also known as \emph{tensor product}.
Given two tensors~$\A$ and~$\B$ of orders~$P$ and~$Q$ respectively, the tensor (outer) product~$\A\otimes\B$ is the tensor of order~$P{+}Q$ defined by:
\be
(\A\otimes\B)_{d_1{\ldots}d_{P{+}Q}}=\A_{d_1{\ldots}d_P}\cdot\B_{d_{P{+}1}{\ldots}d_{P{+}Q}}
\label{eq:tprod}
\ee
The multiplication in the definition of the tensor product is suitable for convolutional arithmetic circuits (linear activation, product pooling), but not for other models.
However, by replacing multiplication with a different operator~$g:\R\times\R\to\R$, we may extend the equivalence to other types of convolutional networks, and in particular, to convolutional rectifier networks.
The next paragraph provides details.

Consider a convolutional arithmetic circuit, \ie~an instance of the architecture depicted in fig.~\ref{fig:base_cac}.
As we have seen in sec.~\ref{sec:cac_htd}, this network corresponds to some hierarchical tensor decomposition~$\mathbf{D}$.
Suppose now that we modify the network by adding point-wise activations~$\sigma(\cdot)$ after each convolution, and replacing product pooling with a different pooling operator~$P\{\cdot\}$.\footnote{
For example, $\sigma(\cdot)$~can be chosen as ReLU ($\sigma(z){=}\max\{z,0\}$), while~$P\{\cdot\}$ could be set to max ($P\{c_j\}{=}\max\{c_j\}$).
}
Define the \emph{activation-pooling operator}:
\be
g:\R\times\R\to\R~~,~~g(a,b)=P\{\sigma(a),\sigma(b)\}
\label{eq:act_pool_op}
\ee
and consider the \emph{generalized tensor product}~$\otimesg$ obtained by placing~$g(\cdot)$ instead of multiplication in the tensor product~$\otimes$ (eq.~\ref{eq:tprod}):
$$(\A\otimesg\B)_{d_1{\ldots}d_{P{+}Q}}=g(\A_{d_1{\ldots}d_P},\B_{d_{P{+}1}{\ldots}d_{P{+}Q}})$$
If we replace all instances of the tensor product~$\otimes$ in the decomposition~$\mathbf{D}$ by the generalized tensor product~$\otimesg$, we obtain what is called a \emph{generalized hierarchical tensor decomposition}, naturally denoted by~$\mathbf{D}_g$.
As it turns out, grid tensors of functions realized by the convolutional network with activation~$\sigma(\cdot)$ and pooling~$P\{\cdot\}$, are precisely the tensors represented by the generalized decomposition~$\mathbf{D}_g$ (which is based on the activation-pooling operator~--~eq.~\ref{eq:act_pool_op}).
We thus have a framework for analyzing general convolutional networks, not just convolutional arithmetic circuits.

\medskip

Focusing on the particular case of convolutional rectifier networks (corresponding to the choices $\sigma(z){=}\max\{z,0\}$ and $P\{c_j\}{=}\max\{c_j\}$ or $P\{c_j\}{=}\mean\{c_j\}$), we follow a path similar to that taken in the analysis of convolutional arithmetic circuits (studying ranks of matricized grid tensors), and derive the following claim (see~\cite{\crngtd} for proof):
\begin{claim} \label{claim:crn_eff_depth_exist}
There exist functions realizable by a deep convolutional rectifier network that can only be replicated by a shallow network if that has an exponential number of hidden channels.
\end{claim}
Taking into account the fact that a deep network easily replicates functions of a shallow one (its second to last hidden convolutions can realize the identity mapping), we conclude from claim~\ref{claim:crn_eff_depth_exist} that with convolutional rectifier networks exponential expressive efficiency of depth takes place.
However, \emph{\textbf{unlike in the case of convolutional arithmetic circuits, where the efficiency of depth is complete, with convolutional rectifier networks it is not}}.
This is stated in the following claim (proven in~\cite{\crngtd}):
\begin{claim} \label{claim:crn_eff_depth_incomplete}
A non-negligible (positive measure) set of the functions realizable by a deep convolutional rectifier network can be replicated by a shallow network with just a few hidden channels.
\end{claim}

The expressive efficiency of depth is believed to be the key factor behind the success of convolutional networks (and deep learning in general).
Our analyses indicate that from this perspective, the widely used convolutional rectifier networks are inferior to convolutional arithmetic circuits.\footnote{
One may argue that this does not carry any information allowing a comparison between the two architectures.
Indeed, we have seen that the expressive efficiency of deep convolutional arithmetic circuits \wrt~shallow convolutional arithmetic circuits is complete, whereas that of deep convolutional rectifier networks \wrt~shallow convolutional rectifier networks is incomplete.
A-priori, it may be that depth is more beneficial with convolutional arithmetic circuits, while overall, convolutional rectifier networks are strictly superior in terms of expressiveness.
Apparently, this is not the case~--~it is shown in~\cite{\crngtd} that the expressive efficiency of deep convolutional arithmetic circuits is complete not only \wrt~shallow convolutional arithmetic circuits, but also \wrt~shallow convolutional rectifier networks.
Analogously, it is shown that the expressive efficiency of deep convolutional rectifier networks is \emph{incomplete} \wrt~shallow models of both architectures.
}
This leads us to believe that convolutional arithmetic circuits bear the potential to improve the performance of convolutional networks beyond what is witnessed today.
Of course, a practical machine learning model is measured not only by its expressiveness, but also by our ability to train it.
Over the years, massive amounts of research have been devoted to training convolutional rectifier networks.
Convolutional arithmetic circuits on the other hand received far less attention, although they have been successfully trained in recent works, showing promising results in different settings (\cf~\cite{\deepsimnets,\tmm}).
We believe that developing effective methods for training convolutional arithmetic circuits, thereby fulfilling their expressive potential, may give rise to a deep learning architecture that is provably superior to convolutional rectifier networks but has so far been largely overlooked.

\section[Inductive Bias of Pooling Geometry]{Inductive Bias of Pooling Geometry \textit{(\cite{\inductive})}} \label{sec:ind_geo}

In this section we focus on question~\ref{question:ind_geo} from sec.~\ref{sec:eff_ind:qs}.
Specifically, we study the effect of a convolutional network's pooling geometry on its ability to model interactions among regions of its input.

Let~$f(x_1,x_2,\ldots,x_N)$ be a function realized by a convolutional network, where~$x_1{\ldots}x_N$ are input elements, for example pixel intensities of a gray-scale image.
Interactions modeled by~$f(\cdot)$ between regions of its input are formalized through the notion of separation rank~--~a commonly used measure in numerical analysis (\cf~\cite{beylkin2002numerical}), which is also deeply rooted in the world of quantum physics (see sec.~\ref{sec:ind_width}).
Let~$(I,J)$ be a partition of input elements, \ie~$I$ and~$J$ are disjoint subsets of~$\{1,2,\ldots,N\}$ whose union covers the entire set.
The \emph{separation rank of~$f(\cdot)$ \wrt~$(I,J)$}, denoted~$sep(f;I,J)$, measures the strength of interaction~$f(\cdot)$ models between the input elements corresponding to~$I$~--~$\{x_i\}_{i{\in}I}$, and those corresponding to~$J$~--~$\{x_j\}_{j{\in}J}$.
Assume for simplicity of notation, and without loss of generality, that~$I=\{1,2,\ldots,K\}$ and~$J=\{K{+}1,K{+}2,\ldots,N\}$.
If~$f(\cdot)$ is separable \wrt~$(I,J)$, meaning there exist functions~$g(\cdot)$ and~$h(\cdot)$ such that:
$$f(x_1,\ldots,x_N)=g(x_1,\ldots,x_K) \cdot h(x_{K{+}1},\ldots,x_N)$$
then under~$f(\cdot)$, there is absolutely no interaction between~$\{x_i\}_{i{\in}I}$ and~$\{x_j\}_{j{\in}J}$.\footnote{
In a statistical setting, where~$f(\cdot)$ is a probability density function, separability \wrt~$(I,J)$ corresponds to statistical independence between~$\{x_i\}_{i{\in}I}$ and~$\{x_j\}_{j{\in}J}$.
}
In this case, by definition,~$sep(f;I,J)=1$.
If the function~$f(\cdot)$ itself is not separable, but can be written as a sum of two separable functions, then~$sep(f;I,J)=2$.
If~$f(\cdot)$ cannot be written as a sum of two separable functions but can be expressed as a sum of three separable functions then~$sep(f;I,J)=3$, and so forth.
In general, the higher~$sep(f;I,J)$ is, the farther~$f(\cdot)$ is from separability \wrt~$(I,J)$, \ie~the stronger the interaction it models between~$\{x_i\}_{i{\in}I}$ and~$\{x_j\}_{j{\in}J}$.

We will analyze the separation ranks brought forth by convolutional arithmetic circuits.
In particular, we focus on the deep network presented in sec.~\ref{sec:cac_htd} (example~\ref{example:deep_ht}), and study the dependence of its separation ranks on the input partition~$(I,J)$.
The following claim (proven in~\cite{\inductive}) links the network's separation ranks to its grid tensors (defined in sec.~\ref{sec:cac_htd}):
\begin{claim} \label{claim:cac_sep_mat_ranks}
Let~$f(\cdot)$ be a function realized by a convolutional arithmetic circuit, and let~$\A$ be its corresponding grid tensor.
For any input partition~$(I,J)$, $sep(f;I,J)$~--~the separation rank of~$f(\cdot)$ \wrt~$(I,J)$, is equal to~$rank\mat{\A}_{I,J}$~--~the rank of~$\A$ when matricized \wrt~$(I,J)$.
\end{claim}
Claim~\ref{claim:cac_sep_mat_ranks} opens the door to an analysis of separation ranks brought forth by convolutional arithmetic circuits through the hierarchical decompositions that represent their grid tensors (see sec.~\ref{sec:cac_htd}).
In the case of the deep network under consideration, the corresponding hierarchical tensor decomposition is~HT (see example~\ref{example:deep_ht}).
The matricization ranks it gives rise to are characterized in the theorem below (see~\cite{\inductive} for proof):
\begin{theorem} \label{theorem:ht_ranks_general}
The maximal rank of tensors generated by HT decomposition, when matricized \wrt~a partition~$(I,J)$, is exponentially high if~$(I,J)$ meets certain conditions, and polynomial (or even linear) otherwise.
\end{theorem}
Given claim~\ref{claim:cac_sep_mat_ranks}, theorem~\ref{theorem:ht_ranks_general} immediately leads to the following corollary:
\begin{corollary} \label{corollary:cac_sep_rank_ind}
A deep convolutional arithmetic circuit can realize exponentially high separation ranks for certain input partitions, whereas for others, it supports separation ranks that are no more than polynomial (or even linear) in network size.
\end{corollary}

\begin{figure*}
\includegraphics[width=\textwidth]{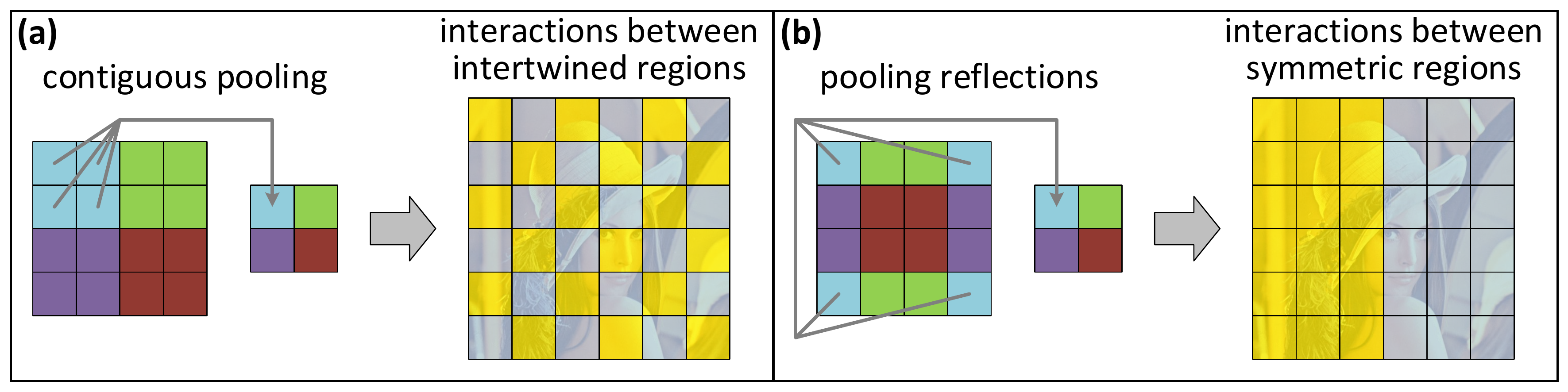}
\vspace{-7mm}
\caption{
Best viewed in color.
Illustrations of different pooling geometries and the type of interactions they favor (see sec.~\ref{sec:ind_geo}).
\textit{\textbf{(a)}}~Standard contiguous windows~--~favor interactions between regions that are highly intertwined, complying with our intuition regarding the statistics of natural images (nearby pixels more correlated than ones that are far apart).
\textit{\textbf{(b)}}~Windows that pool together reflecting elements~--~favor interactions between symmetric regions; may be useful for example when detecting symmetry in medical images.
}
\label{fig:pool_geo}
\vspace{-5mm}
\end{figure*}

Corollary~\ref{corollary:cac_sep_rank_ind} directly relates to inductive bias (see sec.~\ref{sec:eff_ind})~--~it states that \emph{\textbf{a deep network can effectively model strong interactions between some input regions, whereas between others it cannot}}.
Put differently, there are certain, favored interactions that a deep network can model with reasonable size, and on the hand, unfavored interactions that can only be modeled if the network is unfeasibly large.
Apparently, \emph{\textbf{what determines which interactions are favored is the geometry of the network's pooling windows}}.
Standard contiguous windows favor interactions between regions that are highly intertwined (see illustration in fig.~\ref{fig:pool_geo}(a)), reflecting an assumption by which nearby input elements (\eg~image pixels) are more correlated than ones that are far apart.
This explains why the type of pooling geometry most commonly employed in practice is in fact suitable for the kind of data convolutional networks are most frequently applied to (natural images).
More importantly, \emph{\textbf{by modifying pooling geometry one is able to control the type of interactions a network favors, and thereby tailor it to data that departs from the usual domain of natural imagery}} (see illustration in fig.~\ref{fig:pool_geo}(b)).
This is demonstrated empirically in~\cite{\inductive}, with both convolutional arithmetic circuits and convolutional rectifier networks.

\section[Efficiency of Overlapping Operations]{Efficiency of Overlapping Operations \textit{(\cite{\overlaps})}} \label{sec:eff_overlap}

In this section we treat question~\ref{question:eff_overlap} in sec.~\ref{sec:eff_ind:qs}.
Specifically, focusing on the deep network presented in example~\ref{example:deep_ht}, whose hidden convolution and pooling windows do not overlap, we ask whether introduction of overlaps into the latter can lead to expressive efficiency.

\begin{figure*}
\includegraphics[width=\textwidth]{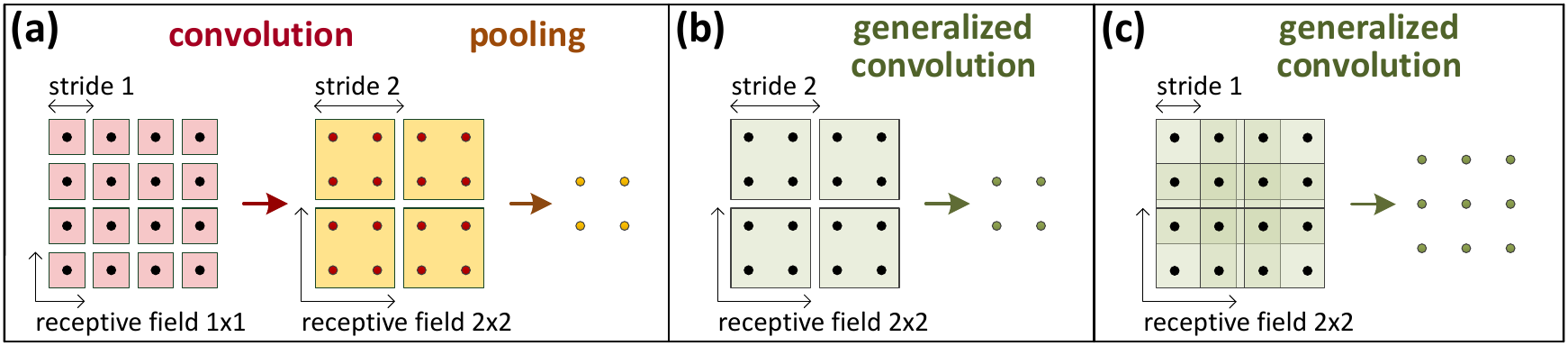}
\vspace{-7mm}
\caption{
Best viewed in color.
Illustrations of receptive fields and strides.
\textit{\textbf{(a)}}~$1{\times}1$~convolution followed by non-overlapping pooling~--~a hidden layer as defined by the baseline convolutional arithmetic circuit architecture (fig.~\ref{fig:base_cac}).
\textit{\textbf{(b)}}~Unification of the convolution-pooling pair in~(a) into a single operation named generalized convolution.
Since the convolution in~(a) is~$1{\times}1$, the receptive fields and strides of the generalized convolution match those of the pooling windows.
\textit{\textbf{(c)}}~Generalized convolution with overlaps~--~obtained by reducing the strides in~(b) to~$1$.
}
\label{fig:overlap}
\vspace{-5mm}
\end{figure*}

Recall from sec.~\ref{sec:cac_htd} that each hidden layer in the deep network consists of convolution followed by pooling, where the convolution has receptive fields~$1{\times1}$, and the pooling decimates feature maps via non-overlapping windows.
We may view the convolution-pooling pair as a single unified operation whose receptive fields and strides match those of the pooling windows (see fig.~\ref{fig:overlap}(a,b)).
This unified operation is referred to as a \emph{generalized convolution}, signifying the fact that it would have been a standard convolution if pooling was based on summations instead of products.
Given the generalized convolution viewpoint, a natural way to introduce overlaps into the network is by reducing strides.
We consider the case of stride~$1$ across all generalized convolutions  (see fig.~\ref{fig:overlap}(c)), and refer to the resulting model as the \emph{overlapping network}.

We would like to show that the overlapping network is expressively efficient \wrt~the original (non-overlapping) one.
In accordance with the definition of expressive efficiency (sec.~\ref{sec:eff_ind}), this calls for establishing two propositions:
\vspace{-2mm}
\begin{enumerate}[(i)]
\item Any function realized by the original network can be replicated by the overlapping one with no more than a linear growth in size
\vspace{-2mm}
\item There exist functions realized by the overlapping network that cannot be replicated by the original one unless the latter's size is allowed to grow super-linearly
\end{enumerate}
\vspace{-2mm}
Proposition~(i) follows from the fact that the overlapping network reduces to the original one if we zero out an appropriately chosen subset of its weights.
For proposition~(ii), we recall that grid tensors of functions realized by the original network are given by HT decomposition (see example~\ref{example:deep_ht}), and that by theorem~\ref{theorem:ht_ranks_general}, there exist partitions~$(I,J)$ under which matricizations of tensors generated by HT decomposition have ranks that are no more than polynomial.
Taken together, these two findings imply that there exist partitions under which matricization ranks of grid tensors realized by the original network are no more than polynomial (in network size).
Denote by~$\mathbf{P}$ the set of such partitions, and consider the following theorem (proven in~\cite{\overlaps}):
\begin{theorem} \label{theorem:overlap_ranks}
There exist partitions in~$\mathbf{P}$ under which matricizations of grid tensors realized by the overlapping network have exponentially high ranks.
\end{theorem}
Theorem~\ref{theorem:overlap_ranks} implies that there exist partitions under which grid tensor matricization ranks are much higher with the overlapping network than they are with the original one.
More precisely, the original network would have to be exponentially large in order to replicate ranks brought forth by the overlapping one.
By this we establish proposition~(ii) above, and prove that \emph{\textbf{a deep convolutional network with overlapping operations can be exponentially expressively efficient \wrt~the same network without overlaps}}.

\section[Inductive Bias of Layer Widths]{Inductive Bias of Layer Widths \textit{(\cite{\dlquantum})}} \label{sec:ind_width}

In this section we address question~\ref{question:ind_width} in sec.~\ref{sec:eff_ind:qs}.
Namely, we study the relation between the width of (number of channels in) each layer in a convolutional network, and the network's ability to model interactions among regions of its input.
Our analysis is based on concepts and tools from the world of quantum physics.

\medskip

A quantum system comprising~$N$ particles is typically represented by a \emph{quantum many-body wave function}, which for our purposes may simply be thought of as a function over~$N$ variables.
A key property of the system, with broad physical implications, is the amount of interaction between different sets of particles.
Interactions are quantified via \emph{quantum entanglement measures} (see~\cite{plenio2007introduction} for an introduction)~--~quantities computed from the many-body wave function.
There are different types of entanglement measures, for example entanglement entropy, geometric measure and \emph{Schmidt number}.
The latter was shown in~\cite{\inductive} to be exactly equivalent to separation rank, as defined in sec.~\ref{sec:ind_geo}.

For simulative purposes, quantum many-body wave functions are usually realized via computational constructs named \emph{tensor networks}.
While an introduction to tensor networks is beyond our scope (the interested reader is referred to~\cite{orus2014practical}), we note here that these can be viewed as graphs in which nodes correspond to tensors (multi-dimensional arrays), edges correspond to tensor modes (axes), and each edge is weighted by the length of its respective mode.
An important class of results relates minimal cuts in the graph underlying a tensor network, to the quantum entanglement measures of many-body wave functions it can realize.
Such results are used by physicists to design tensor networks in accordance with the needs of quantum systems to be modeled.

\medskip

\begin{figure*}
\includegraphics[width=\textwidth]{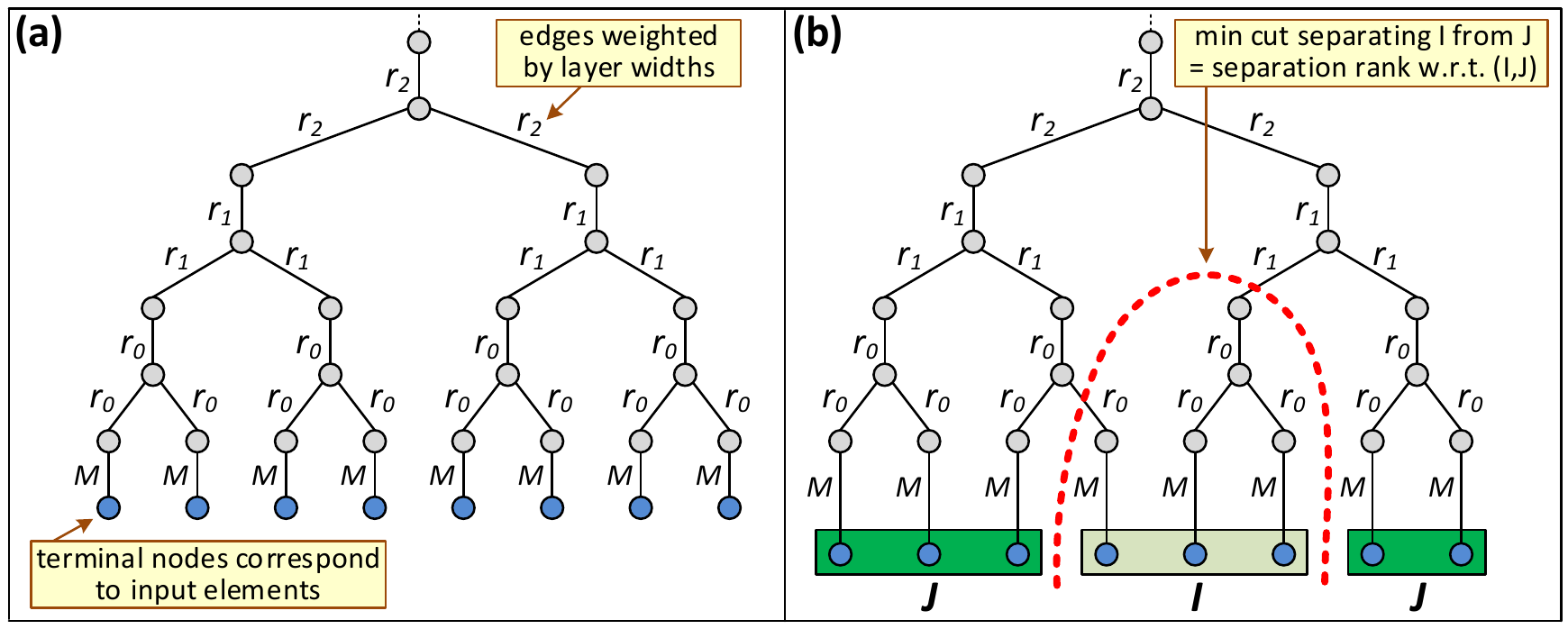}
\vspace{-7mm}
\caption{
Best viewed in color.
\textit{\textbf{(a)}}~Illustration of tensor network (graph) corresponding to the deep convolutional arithmetic circuit presented in example~\ref{example:deep_ht}.
Edges in the graph are weighted by the circuit's layer widths, and a set of terminal (degree-$1$) nodes corresponds to its input elements.
\textit{\textbf{(b)}}~Illustration of the result in theorem~\ref{theorem:sep_rank_min_cut}~--~for any partition of terminal nodes into groups~$I$ and~$J$, the minimal cut separating the two groups is equal to the separation rank \wrt~$(I,J)$ supported by the convolutional arithmetic circuit.
}
\label{fig:tn}
\vspace{-5mm}
\end{figure*}

Returning to the realm of convolutional networks, a clear analogy arises~--~we are also concerned with functions over many local elements (\eg~image pixels or audio samples), and are interested in being able to model the required interactions between them (thereby adhering to the inductive bias~--~see sec.~\ref{sec:eff_ind}).
What opens the door to utilization of tools from quantum physics is the fact that convolutional arithmetic circuits (fig.~\ref{fig:base_cac}) can be cast as tensor networks.
In the tensor network corresponding to a convolutional arithmetic circuit, edges are weighted by layer widths, and there exists a set of terminal (degree-$1$) nodes corresponding to the network's input elements (see fig.~\ref{fig:tn}(a)).

With the connection to quantum physics in place, we rely on the analysis of~\cite{cui2016quantum}, and derive a result characterizing separation ranks (Schmidt entanglements) of a convolutional arithmetic circuit in terms of minimal cuts in its corresponding tensor network (see~\cite{\dlquantum} for proof):
\begin{theorem}
\label{theorem:sep_rank_min_cut}
Let~$\CC$ be a convolutional arithmetic circuit, and let~$\Theta$ be its corresponding tensor network.
For any input partition~$(I,J)$, the highest separation rank that may be realized by~$\CC$ \wrt~$(I,J)$, is equal to the minimal (multiplicative) cut in~$\Theta$ separating terminal nodes of~$I$ from those of~$J$ (see illustration in fig.~\ref{fig:tn}(b)).
\end{theorem}
Taking into account the fact that edges in~$\Theta$ are weighted by widths of (number of channels in) layers in~$\CC$, theorem~\ref{theorem:sep_rank_min_cut} can be used to tailor layer widths so as to optimize separation ranks (interactions) of interest.
Namely, given a convolutional arithmetic circuit with a fixed computational budget, an effective approach for distributing layer widths across the network is to maximize minimal cuts of input partitions for which we would like to model strong interactions.
We head on in~\cite{\dlquantum} and focus on the deep network presented in example~\ref{example:deep_ht}, showing that \emph{\textbf{widths of deep layers are important for modeling long-range interactions, whereas for short-range interactions, widths of early layers are those that matter}}.
This is demonstrated empirically with convolutional rectifier networks, exemplifying once again that analyses carried out with convolutional arithmetic circuits produce practical conclusions that are applicable to other types of convolutional networks as well.

\section[Efficiency of Interconnectivity]{Efficiency of Interconnectivity \textit{(\cite{\dcnmtd})}} \label{sec:eff_conn}

In this section we treat question~\ref{question:eff_conn} in sec.~\ref{sec:eff_ind:qs}, which concerns the ability of connectivity schemes to introduce expressive efficiency over the classic feed-forward (chain) approach.
As opposed to our previous analyses, in which the general tendency was to consider 2D~convolutional networks operating on images, we focus here on 1D~networks.
Specifically, we treat \emph{dilated convolutional networks} operating on sequences.
Dilated convolutional networks are a family of models gaining increased attention in the deep learning community.
In particular, they form the basis of Google's WaveNet~(\cite{van2016wavenet}) and ByteNet~(\cite{kalchbrenner2016neural}) models, which provide state of the art performance in audio and text processing tasks.

\begin{figure*}
\includegraphics[width=\textwidth]{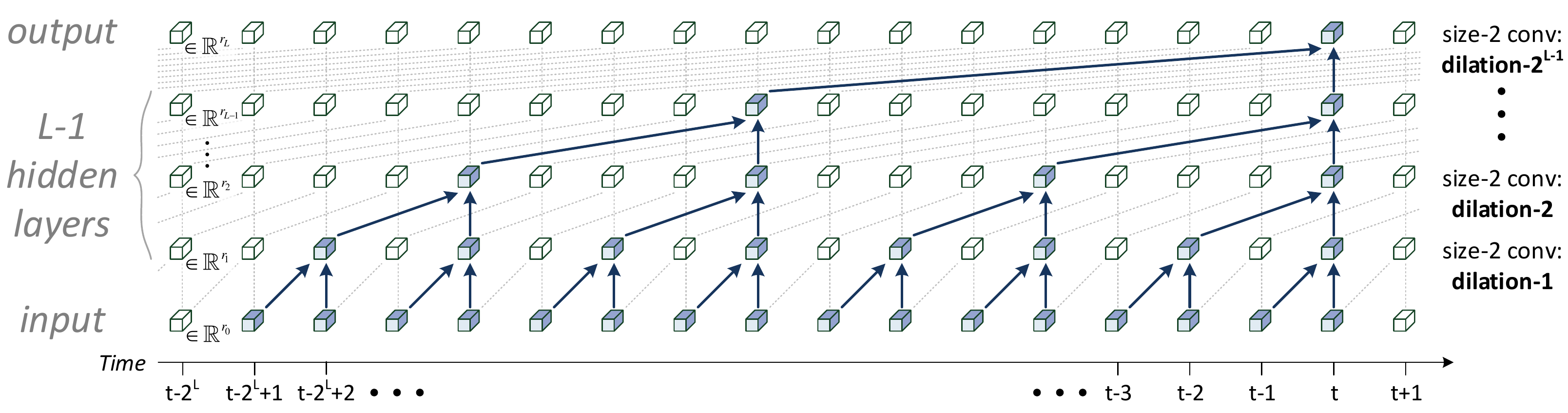}
\vspace{-7mm}
\caption{
Baseline dilated convolutional network~--~underlies Google's WaveNet
}
\label{fig:base_dcn}
\vspace{-2mm}
\end{figure*}

\begin{figure*}
\includegraphics[width=\textwidth]{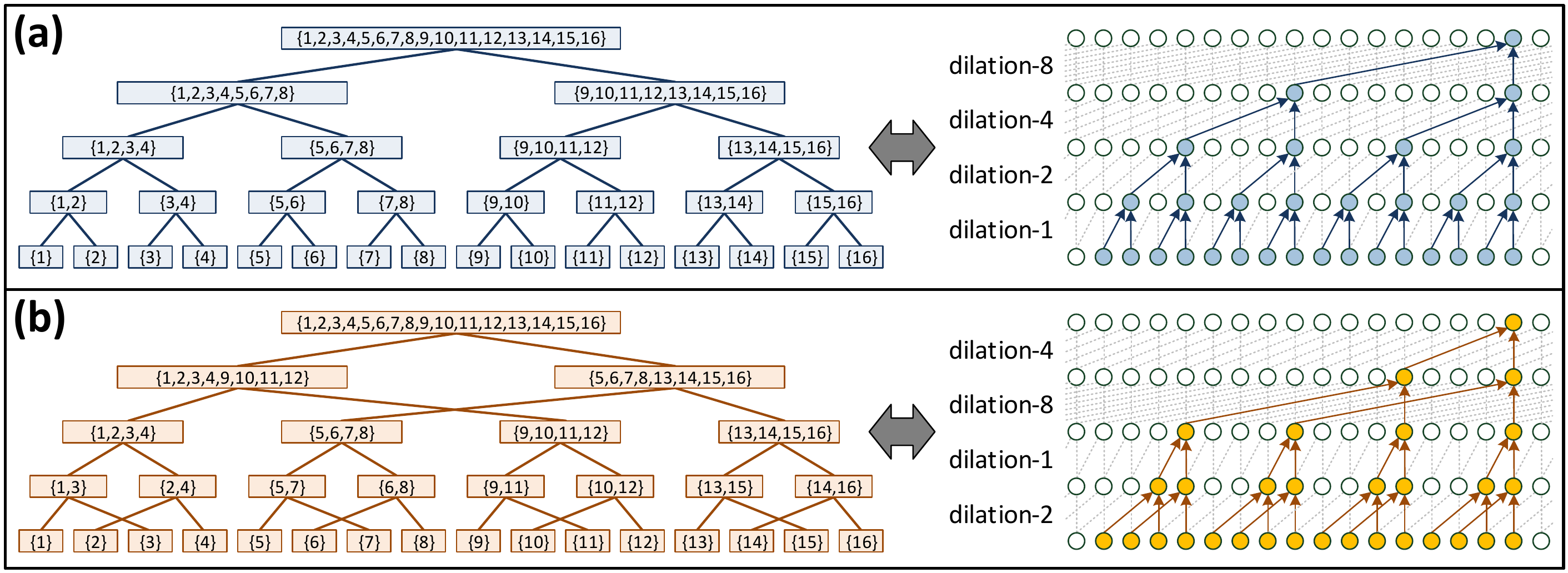}
\vspace{-7mm}
\caption{
Illustrations of trees over tensor modes and the corresponding dilated convolutional networks.
Different trees give rise to different hierarchical tensor decompositions, which represent the grid tensors of networks with different dilations throughout their layers.
\textit{\textbf{(a)}}~Tree corresponding to baseline network (fig.~\ref{fig:base_dcn}).
\textit{\textbf{(b)}}~Alternative tree corresponding to a network with alternative dilations.
}
\label{fig:mode_trees_dilations}
\vspace{-5mm}
\end{figure*}

The dilated convolutional network we consider as baseline is the one underlying WaveNet, depicted in fig.~\ref{fig:base_dcn}.
It is a 1D~convolutional network without pooling, whose convolutional filters are dilated, \ie~incorporate gaps between their elements.
Each layer is characterized by a different dilation, twice as large as that of its preceding layer.
As before, we study functions realized by networks through the hierarchical decompositions that represent their grid tensors (see sec.~\ref{sec:cac_htd}).
In the case of the baseline dilated convolutional network, the hierarchical decomposition adheres to a tree over tensor modes as illustrated in fig.~\ref{fig:mode_trees_dilations}(a).
Modifying the structure of this tree yields a hierarchical decomposition that corresponds to a network with modified dilations throughout its layers~--~see illustration of a particular example in fig.~\ref{fig:mode_trees_dilations}(b).

For the analysis of networks with different connectivity schemes, we introduce the notion of \emph{mixed tensor decompositions}.
Let~$\NN_1$ and~$\NN_2$ be two dilated convolutional networks whose corresponding hierarchical tensor decompositions are based on mode trees~$\T_1$ and~$\T_2$ (respectively).
The mixed tensor decomposition of~$\T_1$ and~$\T_2$ runs their hierarchical decompositions in parallel, while exchanging tensors at different points along the way.
It represents the grid tensors of a mixed network~$\M$, obtained by interconnecting the intermediate layers of~$\NN_1$ and~$\NN_2$.

We would like to show that~$\M$ is expressively efficient \wrt~$\NN_1$ and~$\NN_2$, thereby exemplifying the ability of interconnectivity to introduce efficiency.
As discussed in sec.~\ref{sec:eff_ind}, this requires proving two propositions:
\vspace{-2mm}
\begin{enumerate}[(i)]
\item Any function realized by~$\NN_1$ or~$\NN_2$ can be replicated by~$\M$ with no more than linear growth in size
\vspace{-2mm}
\item There exist functions realized by~$\M$ that cannot be replicated by~$\NN_1$ or~$\NN_2$ unless their size is allowed to grow super-linearly
\end{enumerate}
\vspace{-2mm}
Proposition~(i) follows from the fact that the mixed network~$\M$ reduces to one of the networks it comprises ($\NN_1$~or~$\NN_2$) if we set the opposite network's weights to zero.
For proposition~(ii), we compare matricization ranks under the hierarchical decompositions of~$\T_1$ and~$\T_2$, to those brought forth by their mixture.
This results in the following theorem (see~\cite{\dcnmtd} for proof):
\begin{theorem} \label{theorem:mtd_vs_htd}
Let~$\T_1$ and~$\T_2$ be different trees over tensor modes, and consider the hierarchical decompositions they give rise to.
These decompositions must grow (in terms of the number of intermediate tensors~--~sec.~\ref{sec:cac_htd}) at least quadratically to replicate tensors generated by their mixture.
\end{theorem}
From a network perspective, theorem~\ref{theorem:mtd_vs_htd} translates to:
\begin{corollary} \label{corollary:eff_conn}
Let~$\NN_1$ and~$\NN_2$ be different dilated convolutional networks, and let~$\M$ be a network obtained by interconnecting their intermediate layers.
$\M$~realizes functions that cannot be replicated by~$\NN_1$ or~$\NN_2$ unless these are at least quadratically larger.
\end{corollary}
We conclude that \emph{\textbf{with dilated convolutional networks, interconnectivity brings forth expressive efficiency}}.
Moreover, even a single connection between intermediate layers of different networks already leads to a quadratic gap, which in large-scale settings typically makes the difference between a model that is practical and one that is not.
Empirical evaluation of the analyzed models (carried out in~\cite{\dcnmtd}) demonstrates how adding connections between intermediate layers of different networks improves accuracy, with no additional cost in terms of computation or model capacity.
This serves as yet another indication that in general, expressive efficiency and improved accuracies go hand in hand.

\section{Conclusion} \label{sec:conclusion}

Expressive efficiency and inductive bias are the primary notions for formally reasoning about expressiveness~--~the driving force behind convolutional networks.
Perhaps more important than their role in formalizing common beliefs and explaining empirically observed phenomena, is the potential of expressive efficiency and inductive bias to provide new tools for network design.
Expressive efficiency can be viewed as the enhancement of a network's expressiveness, whereas inductive bias corresponds to making better use of expressive resources given the needs of a task at hand.
Mounting empirical evidence shows time and time again that both procedures directly lead to improved performance (accuracies in particular).

Through an equivalence to hierarchical tensor decompositions, we analyzed the expressive efficiency and inductive bias of various architectural features in convolutional networks.
Specifically, we studied the effects of network depth, layer widths, geometry of pooling windows, overlapping convolutions, and interconnectivity schemes.
The results derived are not only explanatory~--~they provide concrete steps for controlling expressive efficiency and inductive bias.
For example, guidelines are given for setting layer widths and pooling geometries in accordance with input correlations one wishes to model.
We hope the series of works reviewed in this paper will serve as a first step towards extensive use of hierarchical tensor decompositions for more principled convolutional network design.

\newcommand{\acknowledgments}
{\small{This work, as well as all the papers it reviews, were supported by Intel Collaborative Research Institute for Computational Intelligence (ICRI-CI), by ISF Center, and by the European Research Council (TheoryDL project).
Nadav Cohen is supported by a Google Doctoral Fellowship in Machine Learning.}}
\ifdefined\COLT
	\acks{\acknowledgments}
\else
	\ifdefined\CAMREADY
		\subsubsection*{Acknowledgments}
		\acknowledgments
	\fi
\fi

\section*{References}
\small{
\bibliographystyle{plainnat}
\bibliography{refs.bib}
}

\end{document}